\documentclass[letterpaper, 10 pt, conference]{ieeeconf}

\IEEEoverridecommandlockouts

\usepackage{amsmath} 

\usepackage{balance}
\usepackage{amssymb}
\usepackage{amsfonts}
\usepackage{enumerate}
\usepackage{tabulary}
\usepackage{multirow}
\usepackage{url}
\usepackage{cite}
\usepackage{lipsum}  
\usepackage{graphicx}
\usepackage{epstopdf}
\usepackage{filecontents}
\usepackage[misc]{ifsym}
\usepackage{color}
\usepackage{xcolor}
\usepackage{xxcolor}
\usepackage{setspace}
\usepackage[skins]{tcolorbox}
\usepackage{listings}
\usepackage{lipsum}
\usepackage{booktabs}
\usepackage[lined,boxed,commentsnumbered,linesnumbered]{algorithm2e}

\definecolor{codegreen}{rgb}{0,0.6,0}
\definecolor{codepurple}{rgb}{0.58,0,0.82}
\definecolor{backcolour}{rgb}{0.95,0.95,0.92}
\lstdefinestyle{buzz}{
    backgroundcolor=\color{black!5},   
    commentstyle=\color{codegreen},
    keywordstyle=\color{blue},
    numberstyle=\tiny\color{black!30},
    stringstyle=\color{codepurple},
    basicstyle=\footnotesize\ttfamily,
    breakatwhitespace=false,         
    breaklines=true,                 
    captionpos=b,                    
    keepspaces=true,                 
    numbers=left,                    
    numbersep=5pt,                  
    showspaces=false,                
    showstringspaces=false,
    showtabs=false,                  
    tabsize=2,
}
\newcommand{\revision}[1]{\textcolor{black}{#1}}

\DeclareMathOperator*{\argmax}{arg\,max}
\DeclareMathOperator*{\argmin}{arg\,min}

\title{\LARGE \bf
An Adversarial Approach to Private Flocking in Mobile Robot Teams
}

\author{Hehui Zheng$^{1}$, Jacopo Panerati$^{2}$, Giovanni Beltrame$^{2}$, Amanda Prorok$^{1}$\thanks{$^{1}$Hehui Zheng and Amanda Prorok are with the Department of Computer Science and Technology, University of Cambridge, Cambridge, United Kingdom
        {\tt\small \{hz337, asp45\}@cam.ac.uk}}\thanks{$^{2}$Jacopo Panerati and Giovanni Beltrame are with the Department of Software and Computer Engineering, Polytechnique Montr\'eal, Montr\'eal, Qu\'ebec, Canada
        {\tt\small \{jacopo.panerati, giovanni.beltrame\}@polymtl.ca}}}

\begin{document}

\maketitle
\thispagestyle{empty}
\pagestyle{empty}

\begin{abstract}
Privacy is an important facet of defence against adversaries. 
In this letter, we introduce the problem of \emph{private flocking}. We consider a team of mobile robots flocking in the presence of an adversary, who is able to observe all robots' trajectories, and who is interested in identifying the leader. We present a method that generates private flocking controllers that hide the identity of the leader robot.
Our approach towards privacy leverages a data-driven adversarial co-optimization scheme. We design a mechanism that optimizes flocking control parameters, such that leader inference is hindered. As the flocking performance improves, we successively train an adversarial discriminator that tries to infer the identity of the leader robot. To evaluate the performance of our co-optimization scheme, we investigate different classes of reference trajectories. Although it is reasonable to assume that there is an inherent trade-off between flocking performance and privacy, our results demonstrate that we are able to achieve high flocking performance and simultaneously reduce the risk of revealing the leader. 
\end{abstract}

\section{Introduction}
\label{sec:intro}

To date, with the exception of a few recent works (e.g.,~\cite{ prorok2017privacy, li2019coordinated, zhang2019complete}), the topic of \emph{privacy} remains poorly addressed within robotics at large. Yet, privacy can be an important facet of defence against active adversaries for many types of robotics applications. Using privacy as a defence mechanism is particularly relevant for collaborative robot teams, where individual robots assume different roles with varying degrees of specialization. 
As a consequence, specific robots may be critical to securing the system's ability to operate without failure. Our premise is that a robot's motion may reveal sensitive information about its role within the team.
To avoid threats that arise when the roles can be determined by adversaries, we need methods that ensure the anonymity of robots when their motion can be observed.

In this work, we are interested in achieving flocking behavior with \emph{private leaders}, where privacy refers to preventing the inference of the leader's identity, based on observable motion behavior.
Although classical privacy schemes, such as differential privacy, are now increasingly deployed on continuous control problems~\cite{han2018privacy}, they require knowledge of the output distribution of the dynamical system.
Even though it is straightforward to define a basic flocking control scheme, and even sample results from its output distribution through simulation, there are no readily available analytical models that could be used to represent this distribution.

For this reason, we choose an approach towards privacy that leverages data-driven adversarial co-optimization. In specific, we design a mechanism that optimizes flocking control parameters, such that the risk of leader inference is minimized. As the flocking performance improves, we train an adversarial discriminator that tries to infer the identity of the leader. To evaluate the performance of our co-optimization scheme, we investigate three complementary classes of reference trajectories (line, sine, and chevron).

\subsection{Background}
\label{sec:background}

Flocking is a class of formation control algorithms that rely on velocity synchronization and regulation of relative distances within a group of mobile robots~\cite{olfati-saber2006}. 
The aim of formation control is to drive multiple robots to satisfy certain constraints on their physical position, using local or limited information~\cite{oh2015survey}. Movement in formation is crucial for a wide variety of robotics applications including surveillance~\cite{van2008non}, transportation~\cite{bom2005global}, space flight control~\cite{beard2001coordination}, and environmental monitoring~\cite{li2014multi}. The problem varies greatly depending on the desired topology of the formation, and the sensing and communication capabilities available to the robots. Given these varying constraints and the complexity of the problem, a vast number of formation control schemes have been proposed, e.g.,~\cite{olfati-saber2006,ren2008,reynolds1987,desai2001modeling,oh2015survey, fine2013}.

Formation control schemes generally fall into three classes: \emph{(i)} leader-follower schemes, \emph{(ii)} virtual structure schemes, and \emph{(iii)} behavioral formation control schemes.
\emph{Leader-follower} schemes designate one or more robots as leaders and give them access to the desired trajectory of the formation~\cite{desai2001modeling, gu2009}. 
The followers use local information about the leader's position and kinematics to maintain a specified offset that forms the desired formation.

In \emph{virtual structure} schemes, the robot team is considered a single object with a designated trajectory. Each robot then uses this information in addition to local information in order to plan its own motion. These schemes often involve consensus algorithms to drive the robots' states to a common value~\cite{ren2008}.
In particular, virtual leader approaches~\cite{leonard2001virtual,egerstedt2001control, su2009} have the robots agree on the position of a virtual robot, which is then treated as a leader in some leader-follower algorithm.

\emph{Behavioral formation control} schemes assign simple behaviors, such as cohesion and collision avoidance, to individual robots, with the aim of creating an emergent formation~\cite{cucker2007,turgut2008,reynolds1987,balch1998behavior}. A well-known example of this scheme is flocking, which is typically deployed in larger robot teams~\cite{vasarhelyi2018}.
The movement of the team can be directed by a \emph{tacit leader}~\cite{amraii2014explicit}, which is a robot that does not explicitly identify itself as a leader to other robots. Instead, it follows a reference trajectory itself, and thus, due to flock cohesion, causes the group to move with it in the desired direction. 

As evident from the overview above, formation control schemes are vulnerable when they rely on robots with varying degrees of responsibility.
The most obvious case is found in control schemes that leverage explicit or tacit leadership. If the leader is compromised, the entire formation can be led astray or stopped.
While virtual structure approaches do not have designated leaders, it is nevertheless common for only several robots to directly receive trajectory commands, distributing this information amongst the others through local communication. 
This means that specific robots may act as gate-way nodes, receiving control information before other robots do. 
If these key robots can be identified and compromised by an adversary, the mission of the robot team can be easily disrupted.

\subsection{Contributions}
\label{sec:contributions}

There is a dearth of research specifically tackling the problem of \emph{role privacy} for individual robots participating in a formation control scheme.
In this work, we address this gap by introducing the problem of flocking with private leaders. More specifically, we provide the following contributions:
\begin{itemize}
    \item We formulate the problem of flocking in homogeneous mobile robot teams with private leaders.
    \item We develop an adversarial co-optimization method.
    \item We provide a study of the relation between flocking and privacy performance, and analyze the ensuing trade-off. 
\end{itemize}
The following section summarizes related work. Section~\ref{sec:problem} introduces the problem, and Section~\ref{sec:methods} presents our methodology.
In Section~\ref{sec:results}, we provide simulations to exhibit the behavior of the proposed co-optimization scheme, and discuss the impact on the flocking performance. Finally, Section~\ref{sec:conclusions} concludes the article and draws directions for future research.

\section{Related Work}
\label{sec:related}

Although privacy has not traditionally featured as part of robotics research, several works have appeared in this domain in the last few years. In particular, the problem of motion planning and tracking under privacy has been a subject of focused study.
In~\cite{zhang2019complete} and~\cite{okane2009value}, the authors consider the problem of generating a target tracking policy that simultaneously preserves the target's privacy. The work deals with a powerful adversary, who is interested in computing the location of the target, and is assumed to have access to the full history of tracking information. 
Tsiamis et al.~\cite{tsiamis2019}, too, consider a tracking problem.  
A robot is commanded to track a desired trajectory, which is transmitted through a communication channel that can be compromised by eavesdroppers. They design secure communication codes to encode the trajectory information and hide it from the eavesdroppers. 
\revision{In~\cite{bianchin2020} and~\cite{liu2020}, attackers can spoof the sensor readings and control inputs of a robot. The authors demonstrate the existence of undetectable attacks as well as safe trajectories.}

Our work differs from the aforementioned works. Compared to~\cite{tsiamis2019}, we do not focus on the aspect of information transmission. Our adversarial model is similar to the one in~\cite{zhang2019complete}, which assumes an adversary that has access to a history of trajectory information. 
Yet, we consider a distinct problem setting, where \emph{multiple} robots are involved.
In particular, our goal is to provide \emph{role privacy}; thus, we tackle the problem of preventing an adversary from being able to distinguish the role of one robot from that of another.

The issue of role privacy was explored to some extent in prior work~\cite{prorok2016}. That work considers heterogeneous robot teams, and quantifies how easy it is for an adversary to identify the \emph{type} of any robot in the group, based on an observation of the robot group's dynamic state. The framework, however, builds on the theory of differential privacy, and assumes the availability of an output distribution that describes the robot group's dynamic state. As previously mentioned in Section~\ref{sec:intro}, such a method is not easily applied to the case of flocking, where output distributions are hard to model in analytical form, and where the analytical model for the dynamical system is unknown.

Our approach towards privacy leverages an adversarial co-optimization approach\revision{~\cite{goodfellow2014gan}}. The idea of adversarial privacy was recently presented by Huang et al.~\cite{huang2017}, who formulate it as a constrained minimax game between two players. Their method learns the parameters of a privatizer, which is a generative model that creates private data, and an adversarial model, which tries to infer the private variables from the output of the privatizer.
Although the idea of alternating the optimization of privatizer and adversary is common in both our approaches, we apply the method to a completely different domain.

\section{Problem Statement} \label{sec:problem}

Our work has two objectives: 
\emph{(i)} efficient three-dimensional flocking of a team of mobile robots that follows a trajectory only known to a single leader robot; and \emph{(ii)}, privacy of this leadership---that is, making it challenging for an artificial or human adversary to correctly identify the leader of the flock.
We choose to tackle this problem using a co-optimization framework that simultaneously refines the performance of both the robot team $\mathcal{R}$ and an adversarial discriminator $D$.
\revision{In this section, we formalize all the problem components. In Section~\ref{sec:methods}, we detail their implementation.}

\paragraph{Robot team and flocking models} 
We consider a homogeneous robot team $\mathcal{R}$ comprising of $N$ identical robots.
We refer to the position and velocity of each robot $i \in {[1,..,N]}$ as $\mathbf{p_i} \in \mathbb{R}^3$ and $\mathbf{v_i} \in \mathbb{R}^3$, respectively.
All robots share the same limited sensing range $R$ and each robot's neighborhood $\mathcal{N}^R_i$ is defined as the set of all robots within this radius $\mathcal{N}^R_i = \{ j\in \mathcal{R}\ |\ r_{ij} \leq  R\}$---where $r_{ij}$ is the distance between robot $i$ and $j$. 
We assume that each robot is capable of observing the position and velocity of all its neighbors. Thus, the observation vector $o_i$ of robot $i$ at time $t$ can be written as: 
\begin{equation}
	o_i(t) = \{ (\mathbf{p}(t)_j, \mathbf{v}(t)_j) \ |\ j \in \mathcal{N}^R_i(t)\}.
	\label{eq:o}
\end{equation}

Each robot is modelled as a single integrator
with velocity directly set by control $\mathbf{u}_i$.
In its general formulation, the flocking control input is a function of the current state of a robot's neighborhood
and a vector of parameters $\mathbf{c} \in \mathbb{R}^k$:
\begin{equation}
	\dot{\mathbf{p}}_i(t) = \mathbf{v}_i(t) = \mathbf{u}_i(t) = f(o_i(t),\mathbf{c}).
	\label{eq:p}
\end{equation}

The family of the velocity controllers for the robot team leader $l \in {[1,..,N]}$  is a superset of the flocking controllers in~\eqref{eq:p} as it also includes a contribution $g(\cdot)$ based on the leader's absolute position $\mathbf{p}_l$, its intended trajectory $\chi$, and a separate set of parameters $\mathbf{c}_l \in \mathbb{R}^w$ (with $w \geq k$):
\begin{equation}
	\mathbf{v}_l(t) = \mathbf{u}_l(t) = f(o_l(t),\mathbf{c}) + g(\mathbf{p}_l(t), \chi, \mathbf{c}_l).
	\label{eq:v}
\end{equation}
To automate the optimization of the \emph{flocking with leadership} behavior, we also require a formal measure of its \revision{performance loss $\mathcal{F}_{loss}$}, that is a function in the form:
\begin{equation}
	\mathcal{F}_{loss} : \mathbb{R}^{N \times 6 \times q} \longrightarrow \mathbb{R},
	\label{eq:F}
\end{equation}
where $q$ is the number of time steps produced by sampling $\mathbf{p_i}$ and $\mathbf{v_i}$ over $T$ seconds at a fixed frequency $f_{\mathcal{R}}$.

\paragraph{Adversarial discriminator} The discriminator $D$ is an adversarial agent tasked with unveiling the identity of the robot leader $l$ of flock $\mathcal{R}$.
Our assumption is that $D$ has access to observations $o_D$: the positions of the entire team $\mathcal{R}$ for time windows of $W$ seconds, that is, $o_D = \{ \mathbf{p_i}(t)\ |\ i \in [1,..,N] \wedge t \in [t_0,t_0+W] \}, $ where $t_0$ is an arbitrary observation start time.
Here, we assume no noise and uniform time sampling.
Thus, $D$'s role is to solve a multi-class classification problem by attaching to each observation $o_D$ its presumed leader identifier $l_D$.
Given discrete sampling, at fixed frequency $f_D$, a discriminator is any function in the form:
\begin{equation}
	D(o_D) : \mathbb{R}^{N \times 3 \times (f_D \cdot W)} \longrightarrow [0,1]^N.
	\label{eq:D}
\end{equation}
\revision{The co-domain $[0,1]^N$ represents the vectors of likelihoods of each robot being the leader (see Subsection~\ref{sec:discr}).}

\paragraph{Discriminator performance and privacy} 
To formally define privacy as the ability to reduce the performance of an adversarial discriminator, we first define a loss function, returning non-negative penalties for each mislabelled $o_D$:
\begin{equation}
	\mathcal{L}({D}(o_D),y): \mathbb{R}^N \times [1,..,N] \longrightarrow \mathbb{R}^{\geq 0},
	\label{eq:L}
\end{equation}
\revision{where $y$ is the correct leader identifier.}
The \revision{privacy loss $\mathcal{P}_{loss}$} can then be defined \revision{as a function aggregating the losses of $\hat{D}_{\theta}$---a discriminator implementation with parameters $\theta$}:
\begin{equation}
\mathcal{P}_{loss} = h(\mathcal{L}(\hat{D}_{\theta}(\cdot), \cdot)),
	\label{eq:P}
\end{equation}

\paragraph{Problem (adversarial private flocking)} 
given a robot team $\mathcal{R}$ with sensing capabilities as defined in~\eqref{eq:o}, \emph{(i)} \revision{implement $\mathcal{F}_{loss}$, $\mathcal{L}$, $\mathcal{P}_{loss}$, $\hat{D}_{\theta}$}, and \emph{(ii)} design an adversarial co-optimization framework capable of selecting $\mathbf{c}$, $\mathbf{c}_l$ and \revision{$\theta$} such that \revision{$\mathcal{F}_{loss}$} and \revision{$\mathcal{P}_{loss}$} are simultaneously improved\revision{, through adversarial pressure, by playing the minimax game:} \revision{$\argmin_{\mathbf{c}, \mathbf{c}_l} \mathcal{F}_{loss} + \mathcal{P}_{loss}$, $\argmax_{\theta} \mathcal{P}_{loss}$}.

\section{Methods}
\label{sec:methods}

\begin{figure}[tb]
	\centering
		\includegraphics[scale=0.95]{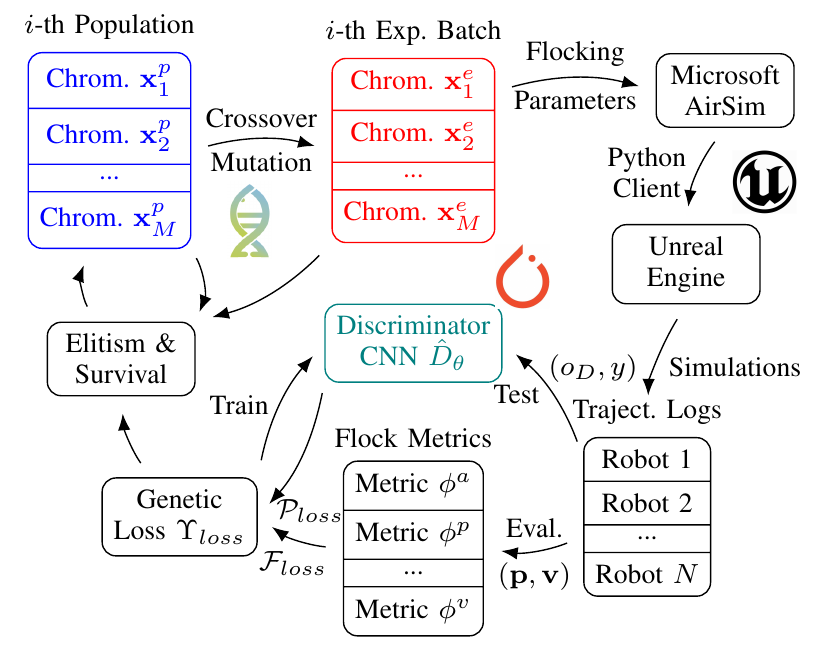} \caption{Block diagram of the proposed implementation of the adversarial co-optimzation approach. From the top left, a complete, clockwise loop presents all the steps in a single generation of genetic optimization.
	}
	\label{fig:2} 
\end{figure}

Our method comprises two components: \emph{(i)} flocking optimization and \emph{(ii)} leader discrimination learning. The former optimizes the controller parameters $\mathbf{c}$, $\mathbf{c}^l$ for more efficient and private  flocking (i.e., minimizing \revision{$\mathcal{F}_{loss}$ and $\mathcal{P}_{loss}$}), while the latter refines \revision{$\hat{D}_{\theta}$} to achieve higher leader-identification accuracy (i.e., increasing \revision{$\mathcal{P}_{loss}$}). We co-optimize these two components in an alternating optimization procedure \revision{(similarly to~\cite{goodfellow2014gan})}, as shown in Figure~\ref{fig:2}. 
\revision{This approach enables \emph{(i)} the successive emergence of behaviors that adapt to adversarial pressure, and \emph{(ii)}, the avoidance of repetitive exhaustive optimization cycles (which can be computationally prohibitive).}
The following text details the sub-components of our co-optimization procedure, which is finally summarized in Subsection~\ref{sec:co-opt}.

\subsection{Flocking Implementation}
\label{sec:flock}

We adopt Reynold's flocking \cite{reynolds1987}, a \revision{decentralized} flocking model consisting of three simple components generating control $\mathbf{u}_i(t) =
\alpha^{\rho} \mathbf{v}^{\rho}_i(t) +
\alpha^{\sigma} \mathbf{v}^{\sigma}_i(t) +
\alpha^{\tau} \mathbf{v}^{\tau}_i(t)$. The first component $\mathbf{v}^{\rho}_i$ 
ensures collision avoidance and it is computed as: \begin{equation}
    \mathbf{v}_i^{\rho} = s\left(\frac{1}{|\mathcal{N}_i^{r^\rho}|}\sum_{j\in \mathcal{N}_i^{r^\rho}}d(r_{\rho}, \mathbf{p}_i, \mathbf{p}_j)\right)
    \label{eq:v-sep}
\end{equation}
with the aid of the helper function $d(\cdot, \cdot, \cdot)$:
\begin{equation*}
d(r^{\rho}, \mathbf{p}_i, \mathbf{p}_j) = 
    \begin{cases}
    \frac{\mathbf{p}_i - \mathbf{p}_j}{\parallel \mathbf{p}_i - \mathbf{p}_j \parallel^2} & \quad 
    {\rm if\ } 0 < \parallel \mathbf{p}_i - \mathbf{p}_j \parallel < r^{\rho} \\
    0 &\quad {\rm otherwise},
    \end{cases}
\end{equation*}
where $s(\cdot)$ is a scale function to transform a position vector into a velocity.
The second component $\mathbf{v}^{\sigma}_i$ is each robot's average neighborhood $\mathcal{N}_i^{r^\sigma}$ velocity:
\begin{equation}
\mathbf{v}^{\sigma}_i =\frac{1}{|\mathcal{N}_i^{r^\sigma}|}\sum_{j\in \mathcal{N}_i^{r^\sigma}}\mathbf{v}_j.
    \label{eq:v-ali}
\end{equation}
To make the flocking cohesive and avoid splitting, 
finally, the third component $\mathbf{v}_i^{\tau}$ leads the robots to move towards the centre of their neighborhood $\mathcal{N}_i^{r_\tau}$:
\begin{equation}
\mathbf{v}_i^{\tau} = s\left(\frac{1}{|\mathcal{N}_i^{r^\tau}|}\sum_{j\in \mathcal{N}_i^{r^\tau}}(\mathbf{p}_j - \mathbf{p}_i)\right).
    \label{eq:v-coh}
\end{equation}

The leader robot's trajectory tracking is implemented by a velocity component $\mathbf{v}_{l}^{\chi}$ which is returned from function $g(\cdot)$ using proportional control in the form:
\begin{equation}
\mathbf{v}_{l}^{\chi}(t) = \omega \frac{(\mathbf{p}_l^{\chi}(t) - \mathbf{p}_l)(t)}{\parallel\mathbf{p}_l^{\chi}(t) - \mathbf{p}_l(t)\parallel}, 
    \label{eq:v-traj}
\end{equation}
where $\omega$ is a control gain
and $\mathbf{p}_l^{\chi}$ is the position on the trajectory at a fixed look-ahead distance.
It is worth noting that the leader robot $l$ can adopt a complete separate set of parameters, meaning that $\alpha^{\rho}$, $\alpha^{\sigma}$, $\alpha^{\tau}$, $r^{\rho}$, $r^{\sigma}$, and $r^{\tau}$ are $\in \mathbf{c}$, while $\alpha^{\rho}$, $\alpha^{\sigma}$, $\alpha^{\tau}$, $r^{\rho}$, $r^{\sigma}$, $r^{\tau}$, and $\omega$ belong to $\mathbf{c}_l$ and can take on different values. The initial position of robot leader $l$ \revision{relative to the rest of} the flock $\mathcal{R}$ is captured by two additional parameters $i_l^x$, $i_l^y$ each $\in [-1,1]$ and included in $\mathbf{c}_l$.
\revision{Note that, while $\mathbf{c}$, $\mathbf{c}_l$ are optimized in a centralized fashion, individual robot control \eqref{eq:p} is still based on local information only.}

\subsection{Flocking Performance Metrics and Genetic Optimization}
\label{sec:metrics}

The \emph{flocking with leadership} model described above incorporates 15 control parameters. Optimizing such a large pool of continuous parameters cannot be done by hand or through parameter sweeping. In our proposal (Figure~\ref{fig:2}), we adopt an evolutionary approach based on classical genetic algorithms (GAs)~\cite{goldberg1988}, that is,
optimization through a biologically-inspired stochastic search. 
Having defined the GA's chromosome $\mathbf{x}$ as the concatenation of $\mathbf{c}$ and $\mathbf{c}_l$, we are left with the non-trivial task of capturing its \revision{loss $\Upsilon_{loss}(\mathbf{x})$}. 

Let $\phi^*(t)$ represent a single performance metric at time $t$. We can define the sample average and variance of metrics $\phi^*$, over simulation time of $T$ samples, as follows:
\begin{equation}
    \overline{\phi^*} = \frac{1}{T}\sum_{i=1}^T\phi^*(t_i) \quad {\rm and} \quad \sigma^* = \frac{1}{T}\sum_{i=1}^T(\phi^*(t_i) - \overline{\phi^*})^2.
    \label{eq:phi}
\end{equation}

To achieve good flock alignment, the literature suggests candidate metrics---velocity correlation \cite{vasarhelyi2018} or polarization \cite{gershenson2016performance}. We chose velocity correlation as it accounts for not only the direction alignment but also the magnitude similarity:
\begin{equation}
\phi^{a}(t) = \frac{1}{N} \sum_{i=1}^N \frac{1}{N-1}\sum_{j=1, j\neq i}^N \frac{\mathbf{v}_i(t) \cdot \mathbf{v}_j(t)}{\parallel\mathbf{v}_i(t)\parallel\cdot \parallel\mathbf{v}_j(t)\parallel}.
    \label{eq:phi-ali}
\end{equation}
This metric should be maximised to encourage alignment within the flock neighborhood \revision{(thus, we consider its negation to compute $\Upsilon_{loss}$)}. 
To ensure dense but also collision-free flocking, we introduce metric $\phi^{r}$:
\begin{equation}
\phi^{r}(t) = \frac{1}{N} \sum_{i=1}^N {\min}(r_{ij})
    \label{eq:phi-r}
\end{equation}
as well as metric $\phi^{p}$:
\begin{equation}
\phi^{p} =
    \begin{cases}
    0 \qquad \qquad {\rm if\ } r_{-}<\overline{\phi^{r}}<r_{+} \\
    {\rm min}(|\overline{\phi^{r}}-r_{-}|, |\overline{\phi^{r}}-r_{+}|)
      \quad 
    {\rm otherwise},
    \end{cases}
    \label{eq:F-r}
\end{equation}
where $[r_{-},r_{+}]$ defines a range of acceptable inter-robot distances. 
The spacing within the flock should also be uniform. Thus, we introduce a metric for the variance of spacing among robots:
\begin{equation}
\phi^{s}(t) = 
\frac{1}{N} \sum_{i=1}^N ({\rm min}(r_{ij}) - \phi^{r}(t))^2.
    \label{eq:S-r}
\end{equation}

To minimize the leader's tracking error to $\chi$, we define:
\begin{equation}
\phi^{\chi}(t) = \parallel\mathbf{p}_l - \mathbf{p}_l^{\chi}\parallel.
    \label{eq:phi-err}
\end{equation}

Finally, to assess the overall flock tracking efficiency, the flock's speed is measured at the centre of mass:
\begin{equation}
\mathbf{v}_{\mathcal{R}} = 
    \frac{1}{N} \left|\sum_{i=1}^N \mathbf{v}_i(t)\right| \label{eq:phi-v}
\end{equation}
and we defined metric $\phi^v$ as:
\begin{equation}
\phi^v =
    \begin{cases}
    0 &\quad{\rm if\ } \overline{\mathbf{v}_{\mathcal{R}}} > \mathbf{v}_{-} \\
    |\overline{\mathbf{v}_{\mathcal{R}}}- \mathbf{v}_{-}|
      &\quad 
    {\rm otherwise},
    \end{cases}
    \label{eq:F-v}
\end{equation}
where $\mathbf{v}_{-}$ is the minimum acceptable flock speed.

We can then combine all these metrics in a vector $\mathbf{m}$:
\begin{equation}
\mathbf{m} = 
    [
    \overline{\phi^{a}},
    \sigma^{a},
\phi^{p},
    \sigma^{r},
    \overline{\phi^{s}},
    \overline{\phi^{\chi}},
    \sigma^{\chi},
    \phi^{v},
    \sigma^{v}
    ],
    \label{eq:m}
\end{equation}
define the hyper-parameter vector $\mathbf{b} \in \mathbb{R}^9$, and write our proposed overall \revision{flocking performance loss as $\mathcal{F}_{loss} = \mathbf{b}^{T} \mathbf{m}$.}
\revision{The GA, in turn, favours better fitness by seeking \emph{smaller} $\Upsilon_{loss}$ values.
Thus, in order to purely optimize for $\mathcal{F}_{loss}$
one can set $\Upsilon_{loss} = \mathcal{F}_{loss}$.}

\subsection{Adversarial Discriminator Design}
\label{sec:discr}

Given the complexity of flocking dynamics and a lack of analytical models that can describe observations of trajectories (as defined in Section~\ref{sec:problem}), we opt to design the discriminator \revision{implementation $\hat{D}_{\theta}$} (Figure~\ref{fig:2}) using a data-driven approach.
Convolutional Neural Networks (CNNs) have proven successful in multi-class classification problem for high-dimensional input with spatial information~\cite{lecun2010}. Their setup closely resembles the problem at hand of the discriminator, which is trained to distinguish the leader robot $l$ from its followers. 
The discriminator's input
$o_D \in \mathbb{R}^{N \times 3 \times (f_D \cdot W)}$ can be directly fed into a CNN as a multi-channel input with shape $N \times 3$ and number of channels $c = f_D \cdot W$.
\revision{The CNN output is a $1 \times N$ vector stating the likelihood of each robot being the leader (see~\eqref{eq:D}).}
We implement~\eqref{eq:P} by first defining $\mathcal{L}$ as the multi-class cross-entropy loss of the predicted output and \revision{$y$, the identifier of the actual leader $l$:}
\begin{equation}
\mathcal{L}(\hat{D}_{\theta}({o_{D}}), {y)} = 
    -{\rm log}\left(\frac{{\rm exp} (\hat{D}_{\theta}({o_{D}})[{y}])}{\sum_{j}{\rm exp}(\hat{D}_{\theta}({o_{D}})[j])}\right),
    \label{eg:L-xe}
\end{equation}
and, finally, privacy $\mathcal{P}$ as:
\begin{equation}
    \mathcal{P} = \frac{1}{\sum_{\mathbf{O}} \mathcal{L}(\hat{D}_{\theta}(\cdot),\cdot) + \gamma }
    \label{eg:P-score}
\end{equation}
where $\gamma$ is a hyper-parameter and $\mathbf{O}$ a set of $(o_D,y)$ pairs.

\subsection{Genetic Optimization with Adversarial Training}
\label{sec:co-opt}
The co-optimization of \revision{flocking $\mathcal{F}_{loss}$ and privacy $\mathcal{P}_{loss}$} is implemented through a loop that alternates genetic optimization generations and training epochs of the CNN, as shown in Figure~\ref{fig:2}.
As the genetic algorithm repeatedly optimizes the controller parameters $\mathbf{c}$, $\mathbf{c}_l$, our framework also refines the \revision{parameters $\theta$ of} discriminator \revision{$\hat{D}_{\theta}$}. The aim is to generate flocking behaviors that fool the discriminator; meanwhile, the trajectories $o_D$ and the correct label $y$ 
of these improved flocking controllers are provided to the discriminator in the next optimization epoch.
Online training is needed to give the discriminator stronger distinguishing ability while the GA continuously tries to beat it.
We achieve this by updating the discriminator network using stochastic gradient descent for one epoch after each GA generation.
To inform the GA about the current performance of \revision{$\hat{D}_{\theta}$}, the \revision{GA's loss} function
is adapted as follows:
\begin{equation}
    \Upsilon_{loss} = 
    \begin{cases}
    \mathcal{F}_{loss} \qquad \qquad {\rm if\ } \mathcal{F}_{loss} \geq \kappa \\
    \beta \cdot \mathcal{F}_{loss} + (1-\beta) \cdot \mathcal{P}_{loss}
      \quad 
    {\rm otherwise}.
    \end{cases}
    \label{eg:fitness}
\end{equation}
Hyper-parameters $\kappa$ and $\beta$ can be used to discard extremely poor flocking performance and tune the trade-off between \revision{$\mathcal{F}_{loss}$ and $\mathcal{P}_{loss}$}, respectively.

\section{Performance Evaluation}
\label{sec:results}

\begin{figure}[tb]
	\centering
		\includegraphics[scale=0.95]{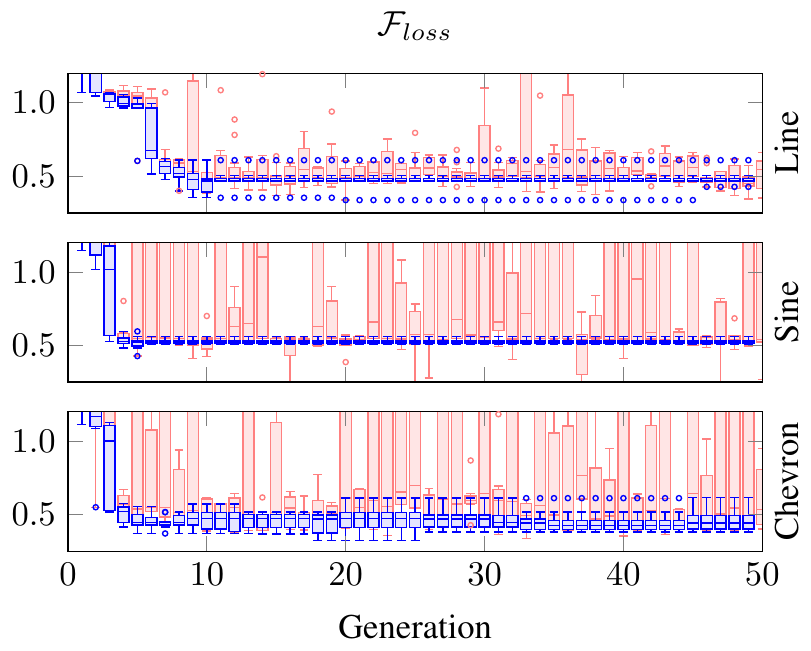} \caption{\revision{Flocking performance loss $\mathcal{F}_{loss}$} through the evolution process for the 3 classes of reference trajectories. The blue box plots represent the distributions of the scores of the chromosomes in the population of the GA. The red box plots represent those of the new experiments of each generation.}
	\label{fig:3} 
\end{figure}

\begin{figure}[tb]
	\centering
		\includegraphics[scale=0.95]{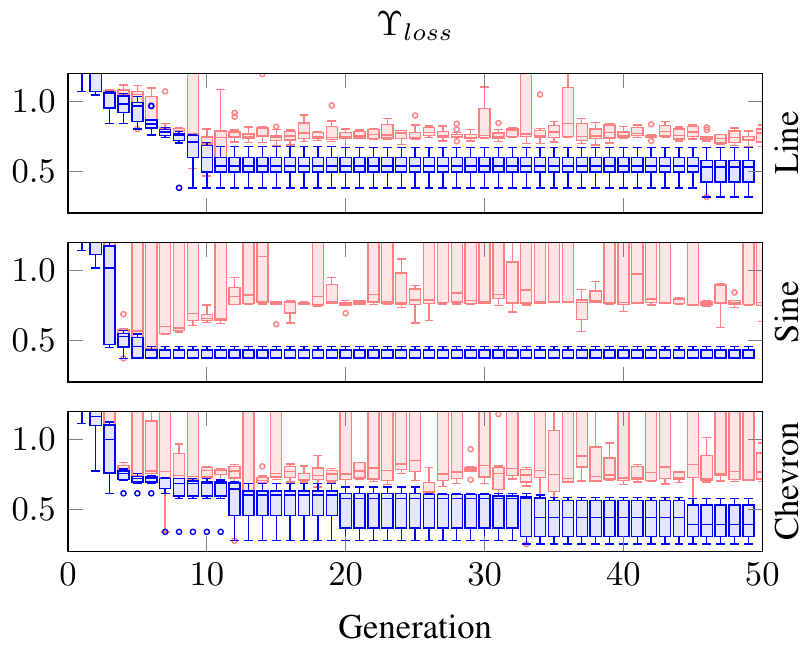} \caption{\revision{Genetic loss $\Upsilon_{loss}$} through the evolution process for the 3 classes of reference trajectories. The blue box plots represent the distributions of the scores of the chromosomes in the population of the GA. The red box plots represent those of the new experiments of each generation.}
	\label{fig:4} 
\end{figure}

\begin{figure}[tb]
	\centering
        \includegraphics[scale=0.95]{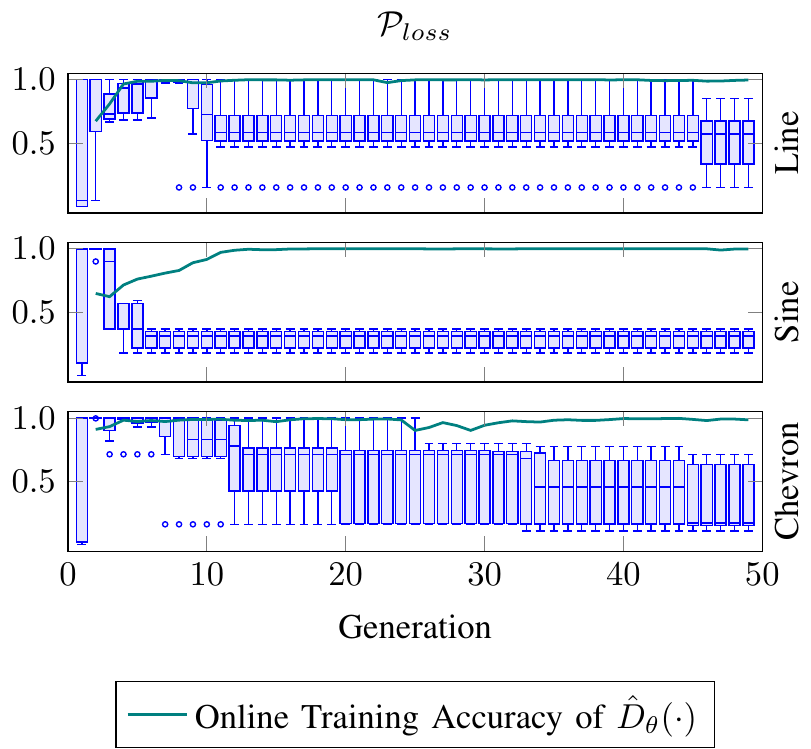} \caption{\revision{Privacy performance loss $\mathcal{P}_{loss}$} through the evolution process for the 3 classes of reference trajectories. The blue box plots represent the distribution of \revision{$\mathcal{P}_{loss}$} for the solutions retained in the population of the genetic algorithm. The teal line represents the training accuracy of \revision{$\hat{D}_{\theta}$}. }
	\label{fig:5} 
\end{figure}

\begin{figure*}[tb]
	\centering
		\includegraphics[scale=0.95]{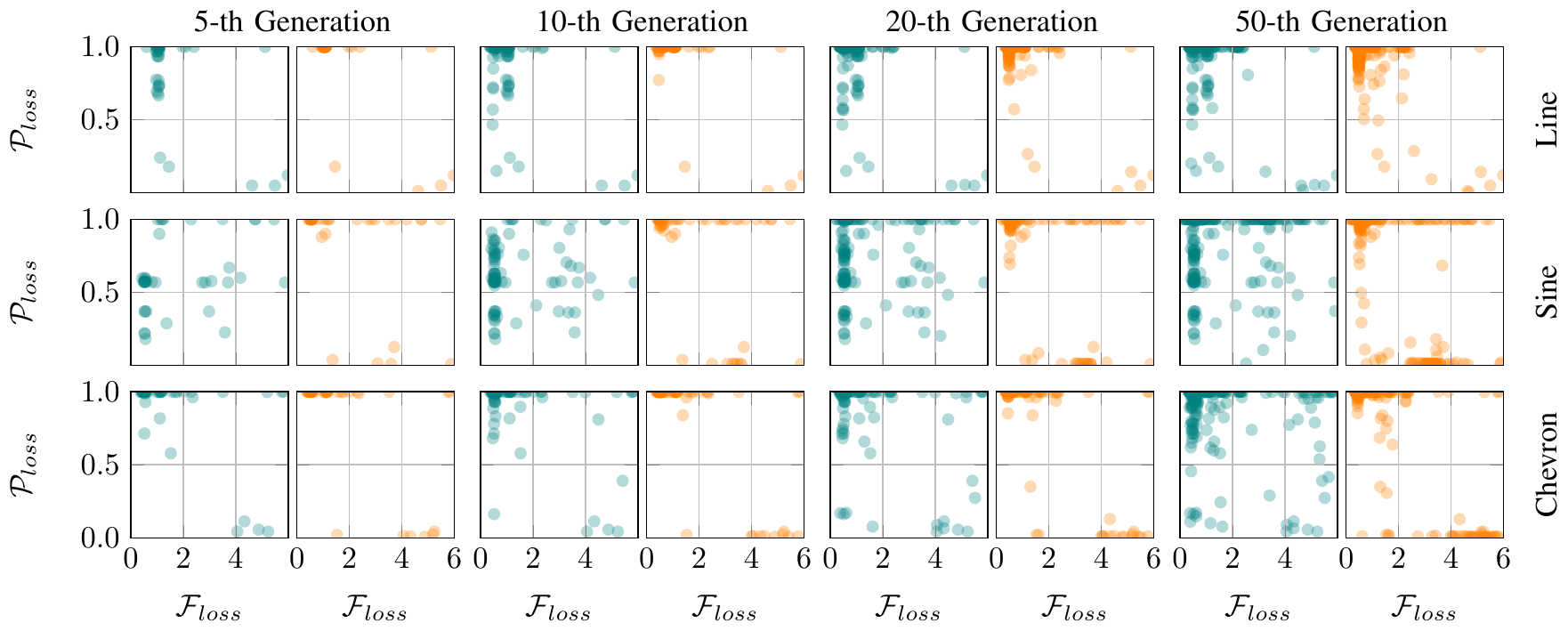} \caption{Trade-off between \revision{flocking performance loss $\mathcal{F}_{loss}$ and privacy loss $\mathcal{P}_{loss}$} for the three classes of leader trajectories (the rows) at four, representative generations (5, 10, 20, and 50, i.e. the columns). The plots with the teal markers present the \revision{privacy loss $\mathcal{P}_{loss}$} returned by the discriminator CNN undergoing the online training described in Subsection~\ref{sec:co-opt}. The plots with the orange markers present the \revision{privacy loss $\mathcal{P}_{loss}$} returned by the originally pre-trained network from Subsection~\ref{sec:setup}. Ideal performance is achieved in the bottom left corner of each plot.}
	\label{fig:0} 
\end{figure*}

\begin{figure*}[tb]
	\centering
		\includegraphics[scale=0.95]{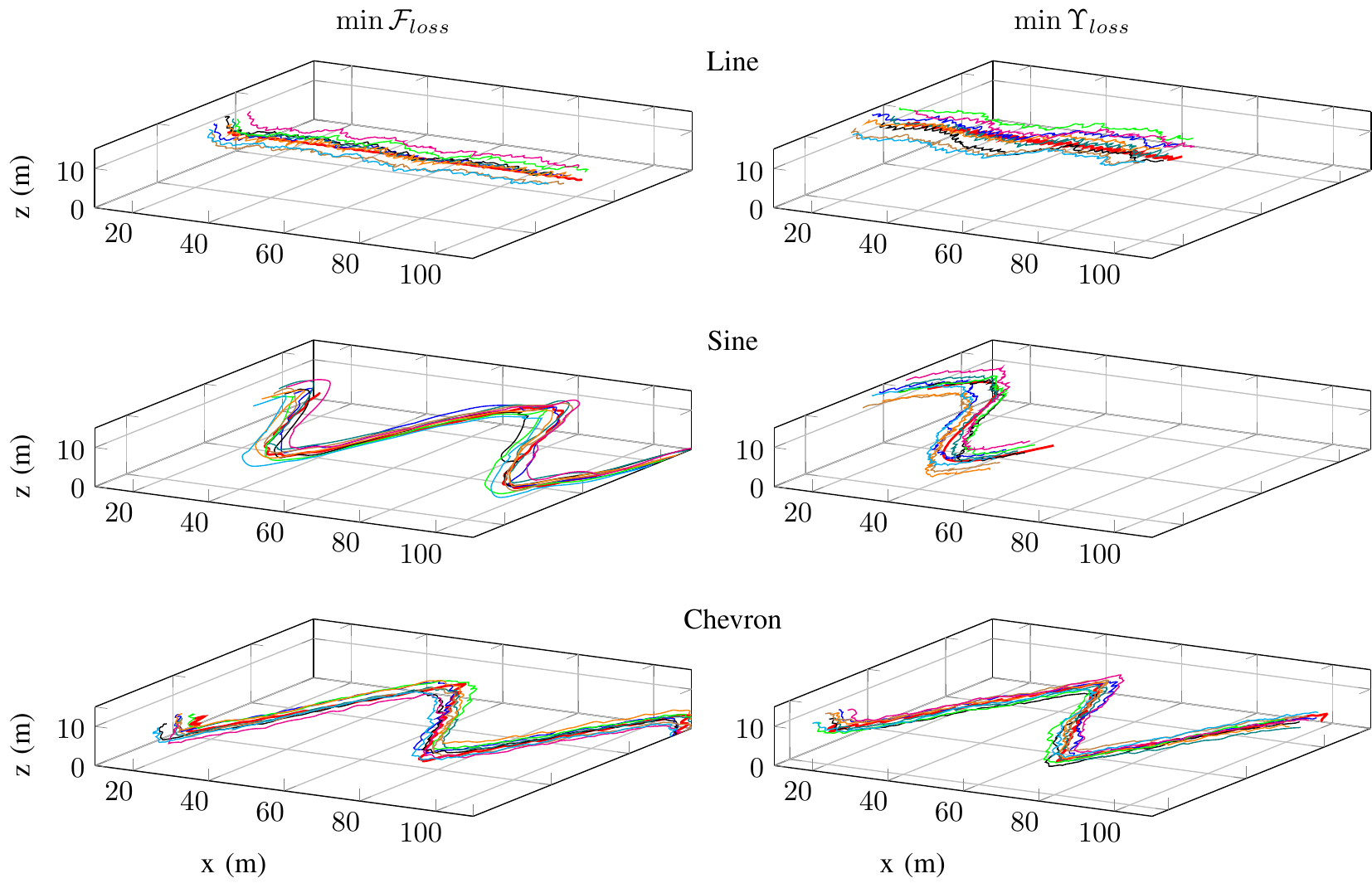} \caption{The genetic champions with respect to \emph{(i)} pure {flocking performance} \revision{$\min \mathcal{F}_{loss}$} (left column) and \emph{(ii)} the joint {privacy/flocking fitness} \revision{$\min \Upsilon_{loss}$} (right column) for the three classes of leader trajectories, over 3-minute simulations. The trajectory trace of the leader robot is drawn in red. Note that amplitude and frequency of the sine and chevron reference trajectory are fixed at design time. \revision{Private flocks, on the sine trajectory in particular, are slower.}
	}
	\label{fig:1} 
\end{figure*}

\begin{figure}[tb]
	\centering
		\includegraphics[scale=0.95]{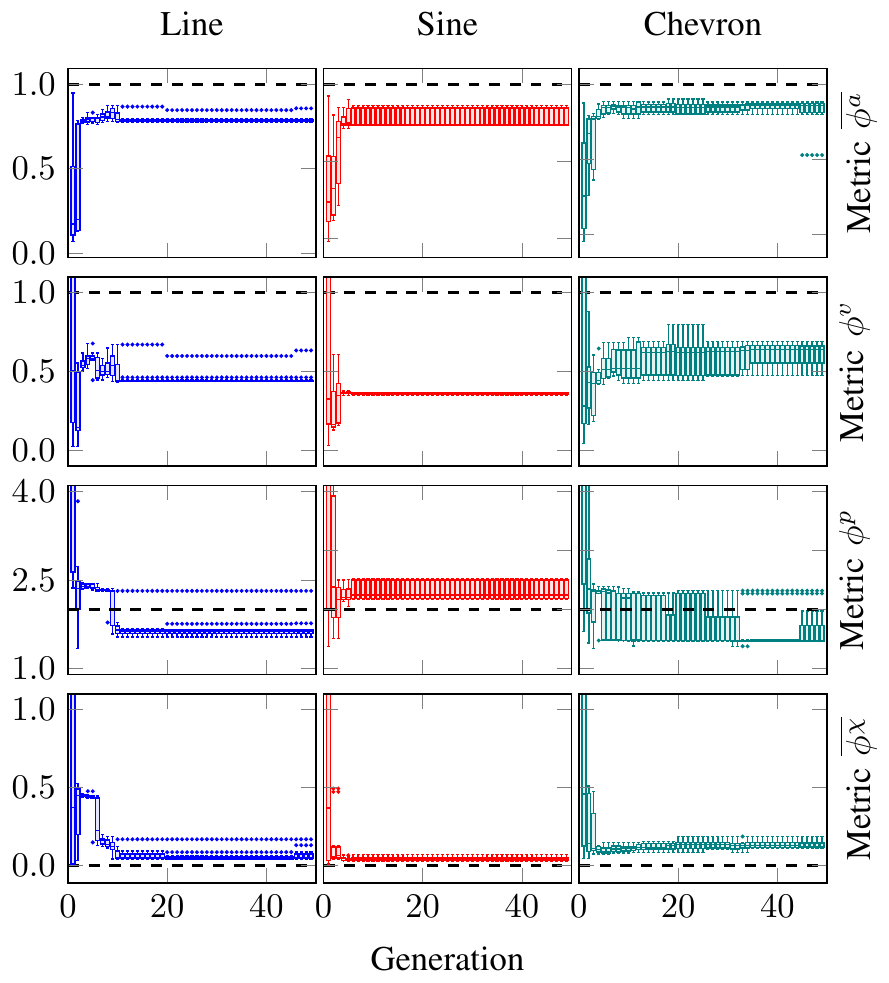} \caption{Evolution of four of the individual components in the \revision{flocking loss $\mathcal{F}_{loss}$} for the three classes of leader trajectories. Each series of box plots represent the distributions of scores of the solutions in the population of the GA. The black dashed line represents the optimization target value.}
	\label{fig:6} 
\end{figure}

\begin{figure}[tb]
	\centering
		\includegraphics[scale=0.95]{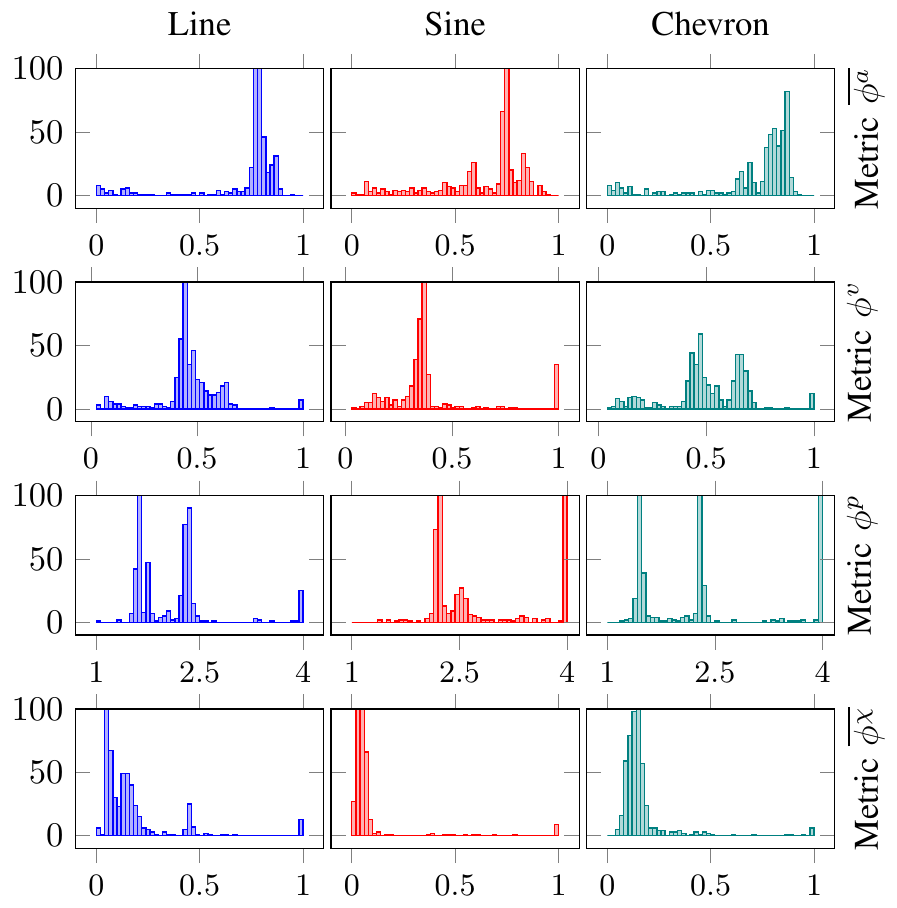} \caption{Distributions over 500 experiments of the same four components of \revision{$\mathcal{F}_{loss}$} from Figure~\ref{fig:6}, for the three classes of leader trajectories.}
	\label{fig:8} 
\end{figure}

In the following, we first describe our simulation setup. We then present the results and discuss our findings.
All of our code is available on GitHub\footnote{\url{https://github.com/proroklab/private_flocking}}.
\revision{Additional results---including the optimized flocking parameters $\mathbf{c}$
and the generalization ability of $\hat{D}_{\theta}$---are also available as supplementary material\footnote{\url{https://arxiv.org/abs/1909.10387}}.}

\subsection{Simulation Setup}
\label{sec:setup}

Our simulations were conducted using a 12-core, 3.2Ghz i7-8700 CPU and an Nvidia GTX 1080Ti GPU with 32 and 11GB of memory, respectively. 
Physics was provided by Unreal Engine release 4.18 (UE4). 
Our flocking control models~\eqref{eq:p} and~\eqref{eq:v} were implemented through the \texttt{AirSim} plugin~\cite{shah2017} and its Python client, which conveniently exposes asynchronous APIs for the velocity control of multiple agents in custom UE4 worlds.
To formalize and run the genetic evolution process, we used ESA's \texttt{pygmo}~\cite{biscani2019} library for massively parallel optimization.
The discriminator's CNN implementation and training used PyTorch~\cite{paszke2017automatic} v1.2.0, and were accelerated with Cuda v10.0 APIs.

The population size of the GA was set to $M = 10$ and evolved over 50 generations. At every generation, \texttt{pygmo} interfaced with \texttt{AirSim} and UE4 to execute 10 3-minute flock flights for each of 10 new chromosomes (experiments)---containing new values for $\mathbf{c}$, $\mathbf{c}_l$. 
Specifically, we used the \texttt{SGA} algorithm with crossover and mutation probabilities of 0.9 and 0.02, respectively, a single crossover point, and elitism set to 3.
Our CNN architecture included a feature extractor {\small \texttt{Conv2d-BatchNorm2d-ReLU-MaxPool2d}} followed by a three-layer {\small \texttt{Linear-ReLU-Linear}} classifier.
In the feature extractor, we used kernels of size 3. Stride and zero-padding were set to 1 for both the {\small \texttt{Conv2d}} and {\small \texttt{MaxPool2d}} layers. The intermediate channel number was set to 16. \revision{The overall number of parameters in $\hat{D}_{\theta}$, $|\theta|$, is 0.23M.
The schematic of $\hat{D}_{\theta}$ is available in our supplementary material$^2$.}

We pre-trained the CNN to give the discriminator reasonable privacy performance at the beginning of the co-optimization process.
We used stochastic gradient descent, for 150 epochs, and 2000 training samples from hand-tuned flocking behaviors following 3 types of leader reference trajectories $\chi \in \{line,\ sine,\ chevron\}$.
The initial leader position ($i_l^x$, $i_l^y$) was uniformly distributed within the flock.
After pre-training, the discriminator achieved 86.9\% test accuracy on hand-tuned flocking.

For the online training of the CNN---i.e., during \texttt{SGA}'s evolution---we used the 100 most recent chromosomes generated by the GA having attained a \revision{flocking loss $\mathcal{F}_{loss} \leq \kappa$}. The learning rate was set to 0.025, and momentum to 0.9.

In the following subsection, we report results for 3 separate evolution processes, one for each of the trajectory types $\in \{line,\ sine,\ chevron\}$, with 9 robots (quadcopters).
These account for 1500 3-minute experiments, that is, over 3 days of simulated flight.

\subsection{Results}
\label{sec:sub-results}

Figure~\ref{fig:3} presents the evolution---over 50 generations---of the \revision{flocking performance loss $\mathcal{F}_{loss}$}
described in Subsection~\ref{sec:metrics}, for the 3 different leader reference trajectories.
As each generation refers to a population of $M$ chromosomes, as well as $M$ new, experimental chromosomes (see Figure~\ref{fig:2}), \revision{$\mathcal{F}_{loss}$} is presented by two series of 50 box plots:
one (blue) describing the evolution of the distribution of \revision{$\mathcal{F}_{loss}$} for the chromosomes in the GA's population; and one (red) describing the evolution of the distribution \revision{$\mathcal{F}_{loss}$} for the experimental chromosomes.
\texttt{SGA} quickly (within 10 generations) and effectively (for all reference trajectories) reaches good values of \revision{$\mathcal{F}_{loss}$}.

The results in Figure~\ref{fig:4} refer to the \revision{GA loss}, \revision{$\Upsilon_{loss}$}~\eqref{eg:fitness}. Similarly to Figure~\ref{fig:3}, Figure~\ref{fig:4} presents information about the evolution of the chromosomes in the GA's population (in blue), and the experimental chromosomes (in red) as series of box plots. 
As \revision{$\Upsilon_{loss}$} depends on \revision{$\mathcal{F}_{loss}$}, we observe similar trends between Figure~\ref{fig:3} and~\ref{fig:4}. However, in the latter, the experimental chromosomes' performance diverges with respect to that of the GA's population, in particular for the $sine$ reference trajectory.
At later stages, we confront the solutions to stronger adversaries, making it more challenging for new solutions to be added to the population.

Figure~\ref{fig:5}, displays a series of box plots capturing the evolution of the distribution of the values of \revision{$\mathcal{P}_{loss}$} for the chromosomes in the population of the genetic algorithm for the 3 reference trajectories, as well as the training accuracy of the CNN \revision{$\hat{D}_{\theta}$}. Notably, improvements in \revision{$\mathcal{P}_{loss}$} are slower than those in \revision{$\mathcal{F}_{loss}$}. We also note that training accuracy of the CNN is slower to converge for the $sine$ reference trajectory.

Figure~\ref{fig:0} shows the trade-off between flocking and privacy performance. The panels in teal show how the genetic evolution successively increases the number of solutions that provide interesting trade-offs between \revision{$\mathcal{F}_{loss}$} and \revision{$\mathcal{P}_{loss}$} (bottom-left is best). 
The pair-wise comparison in each of the four columns of the plot also shows that the \revision{$\mathcal{P}_{loss}$} scores provided by the pre-trained CNN (orange) are more highly polarized towards \emph{non-private}, \emph{good flocking} (top-left) or \emph{private}, \emph{poor flocking} (bottom-right).

In the left and right columns of Figure~\ref{fig:1}, we compare the trajectories generated by the \emph{champion} chromosomes with respect to \revision{$\mathcal{F}_{loss}$} and \revision{$\Upsilon_{loss}$}, respectively, for each of the 3 reference trajectories. The private trajectories (on the right) tend to be slightly slower and less smooth.

Finally, Figures~\ref{fig:6} and~\ref{fig:8} present the evolution (over 50 genetic generations) and the overall distributions (over the 500 experiments in the 50 generations) of four of the flocking performance metrics presented in Subsection~\ref{sec:metrics}: 
\emph{(i)} $\overline{\phi^{a}}$, quantifying the flock's alignment (1 is best);
\emph{(ii)} $\phi^{v}$, quantifying the flock velocity (1 is best);
\emph{(iii)} $\phi^{p}$, quantifying the flock inter-robot spacing (2 is best);
and \emph{(iv)} $\overline{\phi^{\chi}}$, quantifying the leader trajectory tracking error (0 is best). We see that: alignment (row 1) does well for all trajectory classes; flock velocity (row 2) is smaller for the $sine$, and varies more for $chevron$; spacing (row 3) also varies more for $chevron$; and tracking error (row 4) is smallest for $sine$.

\subsection{Discussion}
\label{sec:discussion}
Overall, the results demonstrate that the GA is able to find very good flocking solutions, despite the large parameter space, and despite the increasing strength of the adversarial discriminator. The inclusion of privacy does not appear to harm flocking convergence, yet it is a slower process.

The final performance across the different trajectory classes is comparable, however, \revision{$\Upsilon_{loss}$} converges quicker for the $sine$ reference than for the other two. This is corroborated by the discriminator performance, which shows a slower learning curve for the $sine$. The common denominator among $chevron$ and $line$ is that they are both composed of straight lines. These insights indicate increased difficulty of hiding a leader along linear trajectories.

Although we purposefully constructed a very powerful adversary, it is not a very realistic one. Observations made from a fixed vantage point, or observations that only capture trajectories on a plane, for example, may cause the flock to move in different ways. These avenues remain to be explored in future work.

\section{Conclusions and Future Work}
\label{sec:conclusions}
This work introduced the problem of private flocking and presented a method to generate robot controllers which achieve that feat.
We employed a co-optimization procedure that uses a data-driven adversarial discriminator and a genetic algorithm that optimizes flocking control parameters.
Although we expected an inherent trade-off between flocking performance and privacy, our results demonstrated that we are able to achieve \emph{both} efficient and private flocking, across different classes of reference trajectories. In this work, we considered a worst-case powerful adversary that has access to the complete, non-noisy data. Future work will consider a physical setup with real robots and a realistic adversary, who observes the robot team from specified vantage points, and through potentially noisy sensors.

\section*{Acknowledgements}
\label{sec:acknowledgements}
\small{Hehui Zheng and Amanda Prorok were supported by the Centre for Digital Built Britain, under InnovateUK grant number RG96233, and by the Engineering and Physical Sciences Research Council (grant EP/S015493/1). Their support is gratefully
acknowledged.
Jacopo Panerati was supported by 
a Mitacs Globalink research award and the Natural Sciences and Engineering Council of Canada (NSERC) Strategic Partnership Grant.
The support of Arm is gratefully acknowledged. This article solely reflects the opinions and conclusions of its authors and not Arm or any other Arm entity.

\newpage
\section*{Supplementary Material}
This section contains the supplementary material mentioned in Section~\ref{sec:results} and presented through Figures~\ref{fig:sm:1}, \ref{fig:sm:4}, \ref{fig:sm:5}, \ref{fig:sm:2}, and \ref{fig:sm:3}.

\vspace*{\fill}
\vspace{-3em}
\begin{figure}[ht]
	\centering
		\includegraphics[scale=0.95]{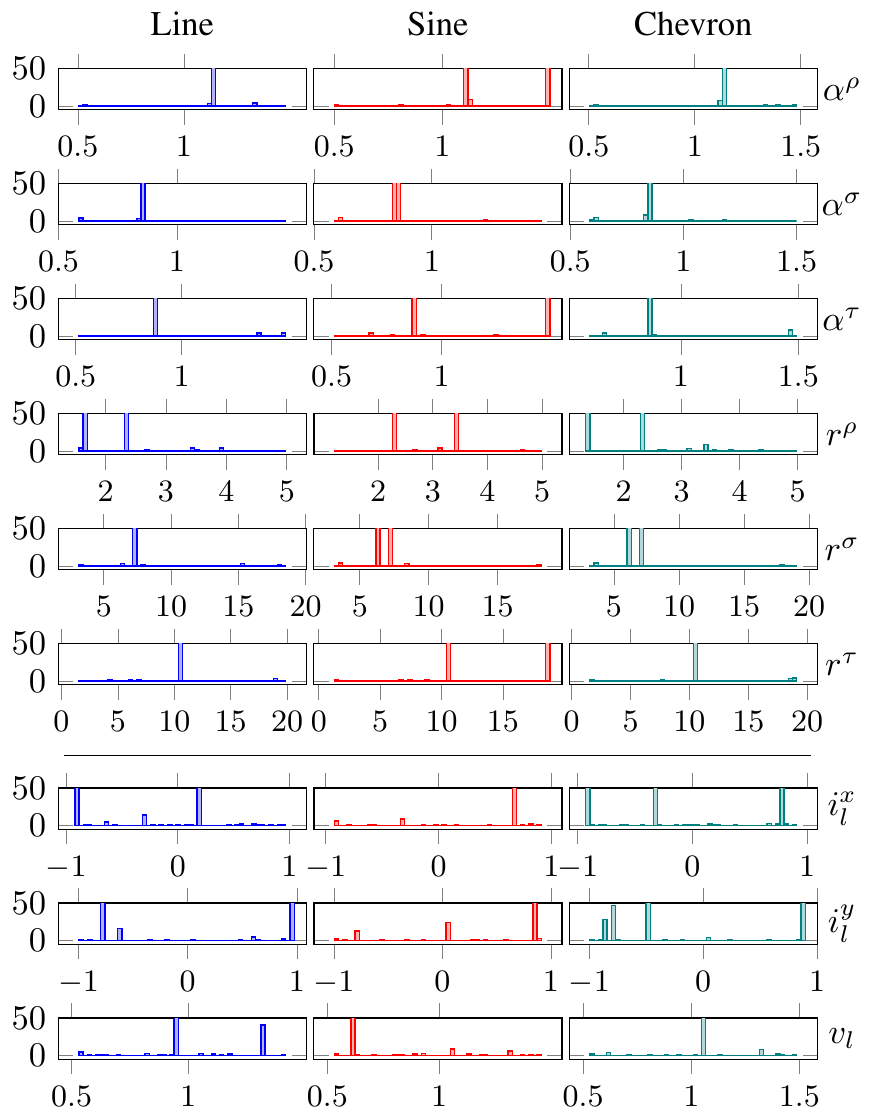} 
	\caption{
Density distributions of the flocking parameters in $\mathbf{c}$, $\mathbf{c}_l$ (see Subsection~\ref{sec:flock}).
The distributions' peaks indicate the most common choices made by the GA throughout the co-optimization process.
The optimal flocking parameters in $\mathbf{c}$---top six rows---are relatively consistent across trajectories; the optimal relative initial position and velocity of the leader---bottom three rows---vary with the choice of reference trajectory.
The preferred leader starting position has a lateral offset ($i^y_l$ close to 1 or -1), especially for line and chevron trajectories.
On the other hand, the co-optimized leader's velocity $v_l$ for the sine trajectory is the slowest.
	}
	\label{fig:sm:1}
\end{figure}
\vspace*{\fill}
\begin{figure}[ht]
	\centering
		\includegraphics[]{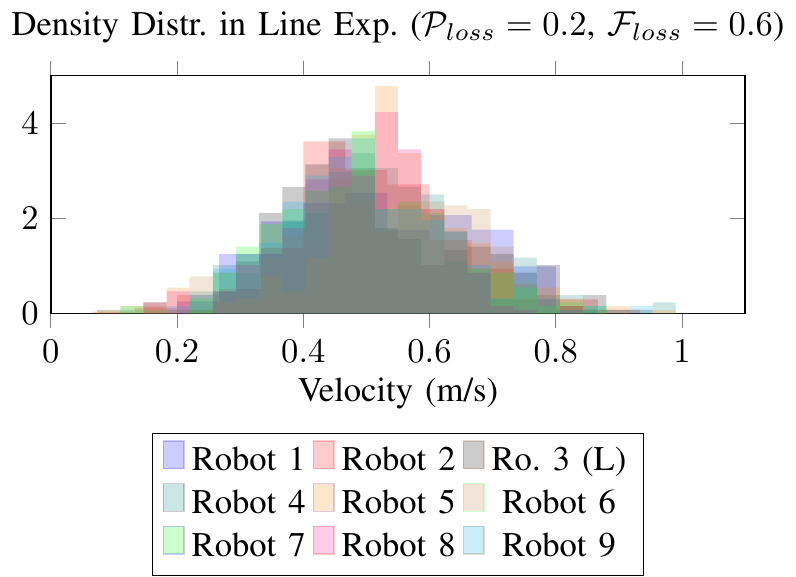} 
		\includegraphics[scale=0.9]{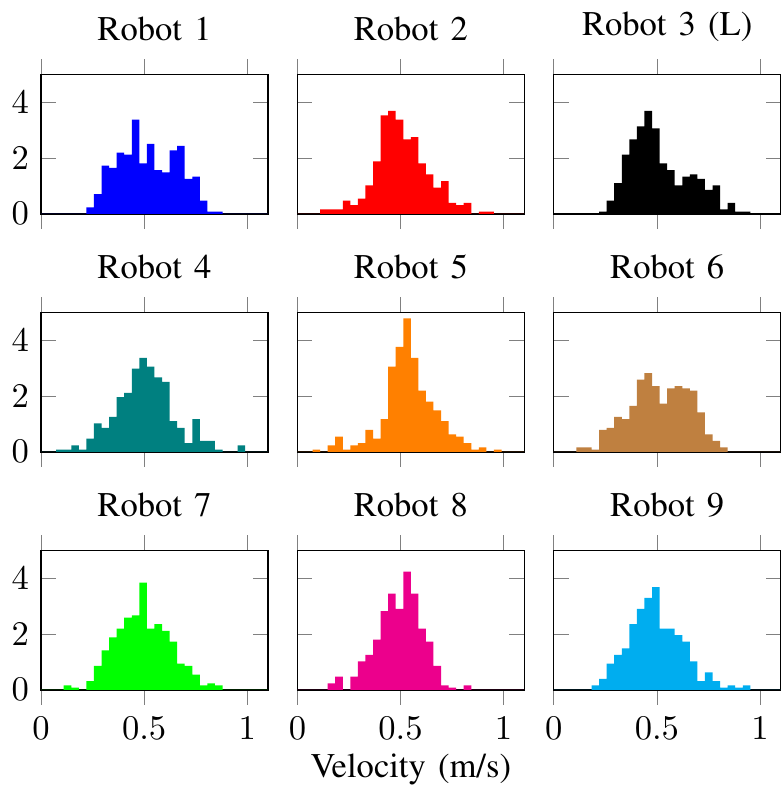} 
	\caption{
Density distributions of the velocities of all robots (followers and leader) in the flock, for a 3' co-optimized line trajectory. We note from the histogram with superimposed distributions (top chart) that the individual distributions are indistinguishable.
Thus, a first-principles-based discriminator would be unable to identify the leader---this difficulty being exacerbated by the fact that only $\sim$3\% of the data shown in the histogram actually goes into each input vector $o_D$ of $\hat{D}_{\theta}$ (i.e., to the observer). 
Even if one could generate trajectories that elicit clear differences between these distributions, the controller's ability to evolve could then learn how to invalidate a fixed, model-based discriminator.
}
	\label{fig:sm:4}
\end{figure}

\begin{figure}[ht]
	\centering
		\includegraphics[width=\columnwidth]{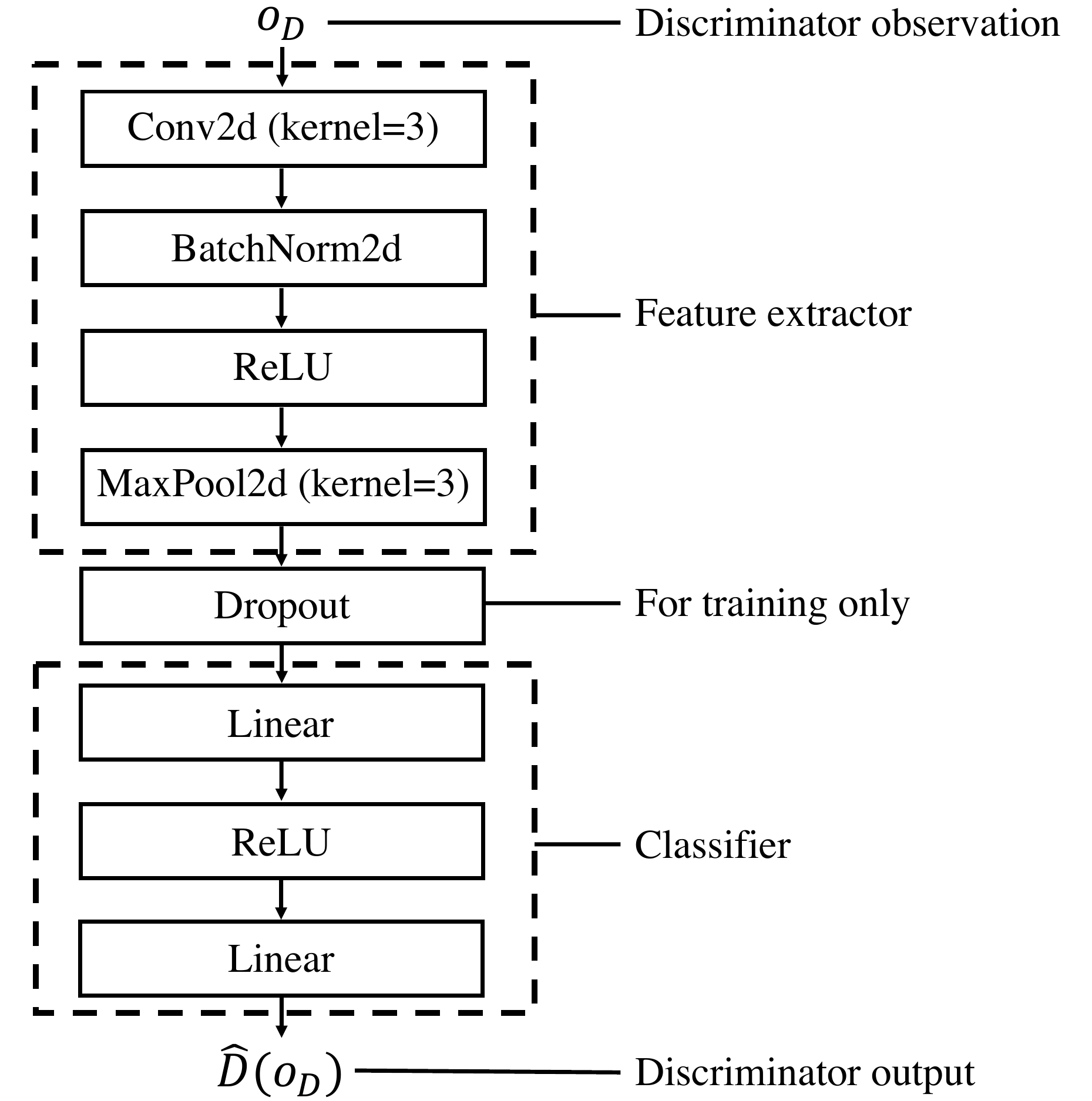} 
	\caption{
This figure presents the schematic of $\hat{D}_{\theta}$'s CNN architecture---as described in Subsection~\ref{sec:setup}.
The third dimension of input $o_D$ (see~\eqref{eq:D}) is a finite number of channels $f_D \cdot W$. In our implementation, we considered $f_D$ of $\sim$2Hz and experimented with $W$ between 1 and 15 seconds, before fixing it to 5''---a value that offered an appropriate trade-off between training accuracy and generalization.
While $o_D$ only contains the robots' positions over time window $W$, we reckon that---in case the robots' velocities were relevant to the discriminator's task---$\hat{D}_{\theta}$ would infer them by simple derivation.
	}
	\label{fig:sm:5}
\vspace{1em}
	\centering
		\includegraphics[scale=0.95]{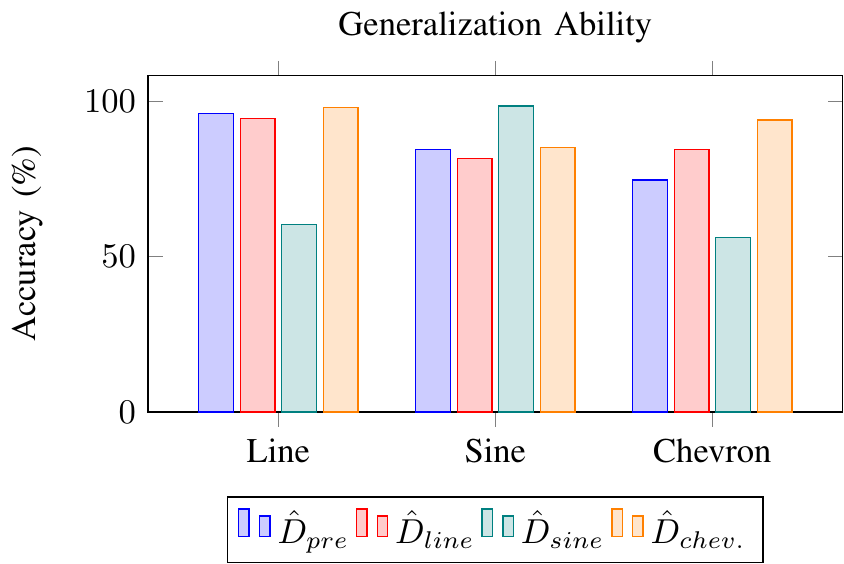} 
	\caption{
We performed additional experiments to evaluate the ability of a CNN trained on specific trajectories to generalize towards others.
$\hat{D}_{pre}$ is pre-trained on a combination of the three trajectories while $\hat{D}_{line}$, $\hat{D}_{sine}$, and $\hat{D}_{chev}$ are trained during co-optimization---each  on a specific trajectory.
We test these four discriminators against optimized, but newly generated, trajectories of all three types.
We observe that the optimized discriminators can achieve on-par or better performance when tested against the specific trajectory they were trained on. While $\hat{D}_{line}$ generalizes well (better than $\hat{D}_{pre}$) to the chevron trajectory and vice versa, $\hat{D}_{sine}$ achieves worse performance than $\hat{D}_{pre}$ against the other two trajectories.
The good generalization between line and chevron trajectories is likely explained by the fact that the typical distance travelled by the flock within the discriminator observation window $W$ is about $3\,$m. Thus, for chevron trajectories composed of $\sim$30-meter segments, the observed data is still mostly straight line-like.
Nonetheless, we note that, since the classification problem is not binary, $\hat{D}_{sine}$ still performs much better than random guessing, even on line and chevron trajectories.
	}
	\label{fig:sm:2}
\end{figure}

\begin{figure}[ht]
	\centering
		\includegraphics[scale=0.95]{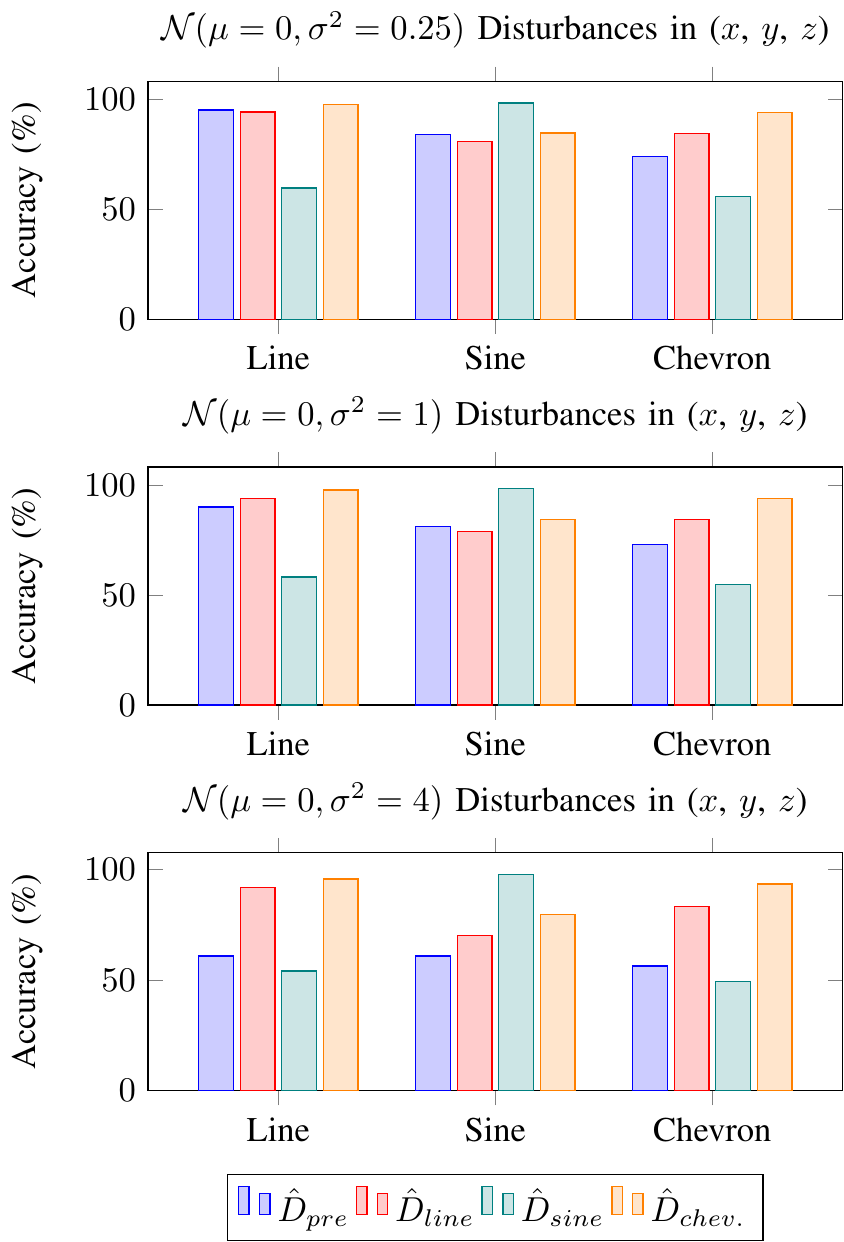} 
	\caption{
We expanded the set of experiments from Figure~\ref{fig:sm:2} to feed our CNNs with test inputs including artificial disturbances.
Normally distributed, zero-mean, fixed-variance disturbances were added in the ($x$, $y$, $z$) observations of the discriminators.
The disturbance variances $[0.25, 1, 4]$ were chosen to resemble the typical travel distance within one observation time window $W$ ($3\,$m) and the average flocking spacing ($1$ to $5\,$m).
For $\sigma^2=0.25$ and $\sigma^2=1$, performance  only slightly degrades, suggesting that the discriminators are, in fact, moderately robust to noise. Even though the discriminators' accuracy decrease for $\sigma^2=4$, when compared to $\hat{D}_{pre}$, we observe that the co-optimized discriminators fare much better.
In fact, the effect of bounded disturbances on the difficulty of leader identification appears to be limited and partially mitigated by the additional training.
	}
	\label{fig:sm:3}
\end{figure}


\begin{thebibliography}{10}
\providecommand{\url}[1]{#1}
\csname url@rmstyle\endcsname
\providecommand{\newblock}{\relax}
\providecommand{\bibinfo}[2]{#2}
\providecommand\BIBentrySTDinterwordspacing{\spaceskip=0pt\relax}
\providecommand\BIBentryALTinterwordstretchfactor{4}
\providecommand\BIBentryALTinterwordspacing{\spaceskip=\fontdimen2\font plus
\BIBentryALTinterwordstretchfactor\fontdimen3\font minus
  \fontdimen4\font\relax}
\providecommand\BIBforeignlanguage[2]{{%
\expandafter\ifx\csname l@#1\endcsname\relax
\typeout{** WARNING: IEEEtran.bst: No hyphenation pattern has been}%
\typeout{** loaded for the language `#1'. Using the pattern for}%
\typeout{** the default language instead.}%
\else
\language=\csname l@#1\endcsname
\fi
#2}}

\bibitem{prorok2017privacy}
A.~Prorok and V.~Kumar, ``Privacy-preserving vehicle assignment for
  mobility-on-demand systems,'' in \emph{IEEE/RSJ International Conference on
  Intelligent Robots and Systems (IROS)}.\hskip 1em plus 0.5em minus
  0.4em\relax IEEE, 2017, pp. 1869--1876.

\bibitem{li2019coordinated}
L.~Li, A.~Bayuelo, L.~Bobadilla, T.~Alam, and D.~A. Shell, ``Coordinated
  multi-robot planning while preserving individual privacy,'' in
  \emph{International Conference on Robotics and Automation (ICRA)}.\hskip 1em
  plus 0.5em minus 0.4em\relax IEEE, 2019, pp. 2188--2194.

\bibitem{zhang2019complete}
Y.~Zhang and D.~A. Shell, ``Complete characterization of a class of
  privacy-preserving tracking problems,'' \emph{The International Journal of
  Robotics Research (IJRR)}, vol.~38, no. 2-3, pp. 299--315, 2019.

\bibitem{han2018privacy}
S.~Han and G.~J. Pappas, ``Privacy in control and dynamical systems,''
  \emph{Annual Review of Control, Robotics, and Autonomous Systems}, vol.~1,
  pp. 309--332, 2018.

\bibitem{olfati-saber2006}
R.~{Olfati-Saber}, ``Flocking for multi-agent dynamic systems: algorithms and
  theory,'' \emph{IEEE Transactions on Automatic Control}, vol.~51, no.~3, pp.
  401--420, March 2006.

\bibitem{oh2015survey}
K.-K. Oh, M.-C. Park, and H.-S. Ahn, ``A survey of multi-agent formation
  control,'' \emph{Automatica}, vol.~53, pp. 424--440, 2015.

\bibitem{van2008non}
D.~Van~der Walle, B.~Fidan, A.~Sutton, C.~Yu, and B.~D. Anderson,
  ``Non-hierarchical uav formation control for surveillance tasks,'' in
  \emph{American Control Conference (ACC)}.\hskip 1em plus 0.5em minus
  0.4em\relax IEEE, 2008, pp. 777--782.

\bibitem{bom2005global}
J.~Bom, B.~Thuilot, F.~Marmoiton, and P.~Martinet, ``A global control strategy
  for urban vehicles platooning relying on nonlinear decoupling laws,'' in
  \emph{IEEE/RSJ International Conference on Intelligent Robots and Systems
  (IROS)}.\hskip 1em plus 0.5em minus 0.4em\relax IEEE, 2005, pp. 2875--2880.

\bibitem{beard2001coordination}
R.~W. Beard, J.~Lawton, and F.~Y. Hadaegh, ``A coordination architecture for
  spacecraft formation control,'' \emph{IEEE Transactions on Control Systems
  Technology}, vol.~9, no.~6, pp. 777--790, 2001.

\bibitem{li2014multi}
S.~Li, Y.~Guo, and B.~Bingham, ``Multi-robot cooperative control for monitoring
  and tracking dynamic plumes,'' in \emph{IEEE International Conference on
  Robotics and Automation (ICRA)}.\hskip 1em plus 0.5em minus 0.4em\relax IEEE,
  2014, pp. 67--73.

\bibitem{ren2008}
W.~Ren and N.~Sorensen, ``Distributed coordination architecture for multi-robot
  formation control,'' \emph{Robotics and Autonomous Systems}, vol.~56, no.~4,
  pp. 324 -- 333, 2008.

\bibitem{reynolds1987}
C.~W. Reynolds, ``Flocks, herds and schools: A distributed behavioral model,''
  in \emph{Proceedings of the 14th Annual Conference on Computer Graphics and
  Interactive Techniques}, ser. SIGGRAPH '87.\hskip 1em plus 0.5em minus
  0.4em\relax New York, NY, USA: ACM, 1987, pp. 25--34.

\bibitem{desai2001modeling}
J.~Desai, J.~Ostrowski, and V.~Kumar, ``Modeling and control of formations of
  nonholonomic mobile robots,'' \emph{IEEE Transactions on Robotics and
  Automation}, vol.~17, no.~6, pp. 905--908, 2001.

\bibitem{fine2013}
B.~T. Fine and D.~A. Shell, ``Unifying microscopic flocking motion models for
  virtual, robotic, and biological flock members,'' \emph{Autonomous Robots},
  vol.~35, no.~2, pp. 195--219, Oct 2013.

\bibitem{gu2009}
D.~{Gu} and Z.~{Wang}, ``Leader?follower flocking: Algorithms and
  experiments,'' \emph{IEEE Transactions on Control Systems Technology},
  vol.~17, no.~5, pp. 1211--1219, Sep. 2009.

\bibitem{leonard2001virtual}
N.~E. Leonard and E.~Fiorelli, ``Virtual leaders, artificial potentials and
  coordinated control of groups,'' in \emph{IEEE Conference on Decision and
  Control}, vol.~3.\hskip 1em plus 0.5em minus 0.4em\relax IEEE, 2001, pp.
  2968--2973.

\bibitem{egerstedt2001control}
M.~{Egerstedt}, X.~{Hu}, and A.~{Stotsky}, ``Control of mobile platforms using
  a virtual vehicle approach,'' \emph{IEEE Transactions on Automatic Control},
  vol.~46, no.~11, pp. 1777--1782, Nov 2001.

\bibitem{su2009}
H.~{Su}, X.~{Wang}, and Z.~{Lin}, ``Flocking of multi-agents with a virtual
  leader,'' \emph{IEEE Transactions on Automatic Control}, vol.~54, no.~2, pp.
  293--307, Feb 2009.

\bibitem{cucker2007}
F.~{Cucker} and S.~{Smale}, ``Emergent behavior in flocks,'' \emph{IEEE
  Transactions on Automatic Control}, vol.~52, no.~5, pp. 852--862, May 2007.

\bibitem{turgut2008}
A.~E. Turgut, H.~{\c{C}}elikkanat, F.~G{\"o}k{\c{c}}e, and E.~{\c{S}}ahin,
  ``Self-organized flocking in mobile robot swarms,'' \emph{Swarm
  Intelligence}, vol.~2, no.~2, pp. 97--120, Dec 2008.

\bibitem{balch1998behavior}
T.~Balch and R.~C. Arkin, ``Behavior-based formation control for multirobot
  teams,'' \emph{IEEE Transactions on Tobotics and Automation}, vol.~14, no.~6,
  pp. 926--939, 1998.

\bibitem{vasarhelyi2018}
G.~V{\'a}s{\'a}rhelyi, C.~Vir{\'a}gh, G.~Somorjai, T.~Nepusz, A.~E. Eiben, and
  T.~Vicsek, ``Optimized flocking of autonomous drones in confined
  environments,'' \emph{Science Robotics}, vol.~3, no.~20, 2018.

\bibitem{amraii2014explicit}
S.~A. Amraii, P.~Walker, M.~Lewis, N.~Chakraborty, and K.~Sycara, ``Explicit
  vs. tacit leadership in influencing the behavior of swarms,'' in \emph{IEEE
  International Conference on Robotics and Automation (ICRA)}.\hskip 1em plus
  0.5em minus 0.4em\relax IEEE, 2014, pp. 2209--2214.

\bibitem{okane2009value}
J.~M. O?kane, ``On the value of ignorance: Balancing tracking and privacy
  using a two-bit sensor,'' in \emph{Algorithmic Foundations of Robotics
  VIII}.\hskip 1em plus 0.5em minus 0.4em\relax Springer, 2009, pp. 235--249.

\bibitem{tsiamis2019}
A.~{Tsiamis}, A.~B. {Alexandru}, and G.~J. {Pappas}, ``Motion planning with
  secrecy,'' in \emph{2019 American Control Conference (ACC)}, July 2019.

\bibitem{bianchin2020}
G.~{Bianchin}, Y.~{Liu}, and F.~{Pasqualetti}, ``Secure navigation of robots in
  adversarial environments,'' \emph{IEEE Control Systems Letters}, vol.~4,
  no.~1, pp. 1--6, Jan 2020.

\bibitem{liu2020}
Y.-C. Liu, G.~Bianchin, and F.~Pasqualetti, ``Secure trajectory planning
  against undetectable spoofing attacks,'' \emph{Automatica}, vol. 112, p.
  108655, 2020.

\bibitem{prorok2016}
A.~Prorok and V.~Kumar, ``A macroscopic privacy model for heterogeneous robot
  swarms,'' in \emph{Swarm Intelligence}, M.~Dorigo, M.~Birattari, X.~Li,
  M.~L{\'o}pez-Ib{\'a}{\~{n}}ez, K.~Ohkura, C.~Pinciroli, and T.~St{\"u}tzle,
  Eds.\hskip 1em plus 0.5em minus 0.4em\relax Cham: Springer International
  Publishing, 2016, pp. 15--27.

\bibitem{goodfellow2014gan}
I.~Goodfellow, J.~Pouget-Abadie, M.~Mirza, B.~Xu, D.~Warde-Farley, S.~Ozair,
  A.~Courville, and Y.~Bengio, ``Generative adversarial nets,'' in
  \emph{Advances in Neural Information Processing Systems 27}, 2014, pp.
  2672--2680.

\bibitem{huang2017}
C.~Huang, P.~Kairouz, X.~Chen, L.~Sankar, and R.~Rajagopal, ``Context-aware
  generative adversarial privacy,'' \emph{CoRR}, vol. abs/1710.09549, 2017.

\bibitem{goldberg1988}
D.~E. Goldberg and J.~H. Holland, ``Genetic algorithms and machine learning,''
  \emph{Machine Learning}, vol.~3, no.~2, pp. 95--99, Oct 1988.

\bibitem{gershenson2016performance}
C.~Gershenson, A.~Muoz-Melndez, and J.~L. Zapotecatl, ``Performance metrics of
  collective coordinated motion in flocks,'' in \emph{Artificial Life
  Conference Proceedings 13}.\hskip 1em plus 0.5em minus 0.4em\relax MIT Press,
  2016, pp. 322--329.

\bibitem{lecun2010}
Y.~{LeCun}, K.~{Kavukcuoglu}, and C.~{Farabet}, ``Convolutional networks and
  applications in vision,'' in \emph{Proceedings of 2010 IEEE International
  Symposium on Circuits and Systems}, May 2010, pp. 253--256.

\bibitem{shah2017}
S.~Shah, D.~Dey, C.~Lovett, and A.~Kapoor, ``Airsim: High-fidelity visual and
  physical simulation for autonomous vehicles,'' in \emph{Field and Service
  Robotics}, 2017.

\bibitem{biscani2019}
\BIBentryALTinterwordspacing
``esa/pagmo2: pagmo 2.11.1,'' Aug. 2019. [Online]. Available:
  \url{https://doi.org/10.5281/zenodo.3364433}
\BIBentrySTDinterwordspacing

\bibitem{paszke2017automatic}
A.~Paszke, S.~Gross, S.~Chintala, G.~Chanan, E.~Yang, Z.~DeVito, Z.~Lin,
  A.~Desmaison, L.~Antiga, and A.~Lerer, ``Automatic differentiation in
  pytorch,'' in \emph{NIPS-W}, 2017.

\end{thebibliography}
\end{document}